\documentclass[conference]{IEEEtran}
\usepackage{cite}
\usepackage{amsmath,amssymb,amsfonts}
\usepackage{algorithmic}
\usepackage{graphicx}
\usepackage{textcomp}
\usepackage{xcolor}
\usepackage{subfig}
\usepackage{float}
\def\BibTeX{{\rm B\kern-.05em{\sc i\kern-.025em b}\kern-.08em
    T\kern-.1667em\lower.7ex\hbox{E}\kern-.125emX}}

\title{Improved Stock Price Movement Classification Using News Articles Based on Embeddings and Label Smoothing}
\author{\IEEEauthorblockN{Luis Villamil\\villamil@cs.fsu.edu} \\ \IEEEauthorblockN{Ryan Bausback\\rwb16@my.fsu.edu} \and \IEEEauthorblockN{Shaeke Salman\\salman@cs.fsu.edu}\\ \IEEEauthorblockN{Ting L. Liu\\tinglliu@bu.edu} \and \IEEEauthorblockN{Conrad Horn\\horn@cs.fsu.edu} \\ \IEEEauthorblockN{Xiuwen Liu\\liux@cs.fsu.edu}}

\begin{document}

\maketitle
\begin{abstract}
Stock price movement prediction is a challenging and essential problem in finance.
While it is well established in modern behavioral finance that the share prices of related stocks often move after the release of news via reactions and overreactions of investors, how to capture the relationships between price movements and news articles via quantitative models is an active area
research; existing models have achieved success with variable degrees. In this paper, we propose to improve stock price movement classification using news articles by incorporating regularization and optimization techniques from deep learning. More specifically, we capture 
the dependencies between news articles and stocks through embeddings and bidirectional recurrent neural networks as in recent models. We further incorporate weight decay, batch normalization, dropout, and label smoothing to improve the generalization of the trained models. To handle high fluctuations of validation accuracy of batch normalization, we propose dual-phase training to realize the improvements reliably. Our experimental results on a commonly used dataset show significant improvements, achieving average accuracy of 80.7\% on the test set, which is more than 10.0\% absolute improvement over existing models. Our ablation studies show batch normalization and label smoothing are most effective, leading to 6.0\% and 3.4\% absolute improvement, respectively on average.
\end{abstract}
\section{Introduction}

One of the principles forming the foundation of modern finance is the Efficient Market Hypothesis (EMH), the theory that, in an efficient market, the stock price fully reflects all available, relevant information \cite{famajstorEMH}. This poses a problem for anyone attempting to make predictions about stock prices since all the information necessary for prediction is already part of the price, so no financial advantage can be reliably produced \cite{TIMMERMANN200415}.

Despite this, it has been shown in several contexts that text-based information such as Twitter \cite{li2017web}, news articles \cite{fung2005predicting}, and even Reddit \cite{bollen2011twitter} could  be used to accurately predict future price movements. For instance, Bollen et al. achieved $87.6\%$ accuracy when predicting up and down movements of stock prices from Twitter \cite{bollen2011twitter}. One has to look no further than the recent incident with ``r/WallStreetBets" on Reddit in order to find an example where sentiment from a text-based internet resource played a significant role in future price movements. In another example, Li et al. \cite{li2017web} described a 2013 fake tweet about Barack Obama being injured, which caused the Dow Jones Industrial Average to drop 100 points within two minutes. Therefore, it is clear the stock market does move in response to media such as news articles. 

However, capturing the complex dependencies between the two systems is a difficult problem. Such a model would need to efficiently compute the relevant news article features in addition to generalizing across all stocks and news articles well. One way to represent those complex dependencies is by using \textit{Natural Language Processing} (NLP) techniques\cite{du-tanaka-ishii-2020-stock}. In recent years, there has been a revolution in such techniques, which in some cases resulted in neural networks outperforming humans at the same task. For instance, NLP has consistently achieved F1 scores as high as $93.2\%$ on the Stanford SQuAD 2.0, a reading comprehension assessment\cite{stanfordSQuAD2}. Humans, on the other hand, only achieved $89.4\%$.

In this paper, we further build upon the work of Du and Tanaka-Ishii \cite{du-tanaka-ishii-2020-stock}, who used such NLP techniques to create embeddings of stock prices and news articles. They then applied those embeddings to the binary classification problem of whether the stock price will move up or down, through the use of a Gated Recurrent Unit (GRU) neural network. We further investigate how deep learning techniques such as batch normalization, dropout, weight decay, and label smoothing improved the performance of a bidirectional GRU (Bi-GRU) neural network with the same settings. First, we describe the classification problem in detail, as well as how the embeddings were generated by Word2Vec and BERT, two NLP techniques. Then we explain the dataset, our new state-of-the-art training method, and the model's performance on that dataset during the experiments. We also briefly discuss the model's performance on the prediction problem. Finally, we address potential problems with the model and its impact, and suggest several improvements for the future.

\section{Related Work}
Stock price movement prediction has become an increasingly studied topic among the NLP researchers working to understand the volatile behavior of financial markets. With the recent development of natural language processing, automatic sentiment analysis has become highly accurate. 

Several studies have developed a number of techniques that tried to predict the stock price movement based on sentiment. For example, Bollen et al. utilized OpinionFinder \cite{wilson2005opinionfinder} to find the positive and negative sentiment of tweets. They also used GPOMS to measure six different human moods from the same tweets in order to predict the closing values from the Dow Jones Industrial Average (DJIA)\cite{bollen2011twitter}. While these type of features can be used to classify sentiment of a text, it is now possible to extract useful information from news articles and other text based content in the form of continuous vectors. More formally, these fixed-length vectors are called embeddings of the texts. These embedding vectors allow numerical computations as they usually contain rich semantic information extracted from the contexts \cite{mikolov2013distributed}.

Word2Vec is an algorithm designed to create trained word embeddings. These embeddings help define the meanings of words by producing a mathematical representation of the word. This representation can then be used to capture relationships with other words using similarity functions like dot product, or cosine similarity. Word2Vec has been done in two algorithms, Skip-gram or Continous Bag of Words (CBOW) \cite{Mikolov2013EfficientEO}. In the Skip-Gram model, it uses a set of sentences (corpus) to predict the neighbors (or contexts) of an input word. In the CBOW method, it predicts the word from the contexts words.
The introduction of Word2Vec enabled new NLP models to be developed for sequences like ELMo, an application of bi-directional LSTM\cite{peters-etal-2018-deep}, using recurrent neural networks (RNN) like Long Short-term memory (LSTM) or Gated Recurrent Units (GRU). LSTM is a type of RNN that uses input and output gates to address the vanishing gradient problem \cite{hochreiter1997long}. GRU is another type of RNN similar to LSTM but with only two gates, a reset gate and update gate, simplifying and speeding up the training \cite{cho2014learning}.

With the recent introduction of transformers, many tasks that were considered challenging have become more feasible. For instance, within the last five years, NLP methods involving transformers have surpassed humans on the precision recall (F1), and exact matching (EM) aspects of reading comprehension questions in the Stanford SQuAD dataset\cite{stanfordSQuAD2}. The majority of methods performing better than humans utilize a transformer system known as BERT.  Bidirectional Encoder Representations from Transformers (BERT)\cite{devlin2019bert}, is a technique that uses transformer models \cite{vaswani2017attention} to classify an entire piece of text instead of word by word. This means that the vectors that are output from BERT have been encoded to represent the context of the sentence. Hence utilizing transformers allows for contextualized representations of sentencing meanings that allow models to learn easier. 

Du and Tanaka-Ishii\cite{du-tanaka-ishii-2020-stock} built from Hu et al.\cite{hu2019listening}, one of the first works that applied the attention mechanism to news-driven stock price movement classification. By utilizing Word2Vec and BERT to create a dual representation of the headlines of articles, Du and Tanaka-Ishii\cite{du-tanaka-ishii-2020-stock} classified the price of a stock in a bi-directional GRU and used a multilayer perceptron for the binary classification. However, they did not explore how deep learning techniques such as batch normalization and label smoothing can further improve generalization. Batch normalization reduces the internal covariance shift of the model through a normalization step that fixes the distribution of layer inputs, thus reducing the dependence of gradients on the scale of the parameters or their initial values \cite{ioffe2015batch}. Label smoothing replaces one-hot encoded (0 or 1) targets with soft targets during calculation of the loss. These soft targets are a linear combination the original labels with a uniform distribution\cite{muller2019does}, and were first shown by Szegedy et al. \cite{szegedy2016rethinking} to significantly improve model generalization. We therefore seek to investigate how these popular and previously effective deep-learning techniques impact the work of Du et al. 

\section{The Stock Price Movement Classification Model}
This section will explain the price movement classification problem, briefly explain how the NLP techniques were used to create the news article embeddings, and give an overview of the proposed model.
\subsection{Price Movement Classification Problem}
For the closed price $p_{t}$ on day $t$ where $t \in \{1,2,...,T\}$, 
and $T$ is the number of trading days in the considered time period \cite{du-tanaka-ishii-2020-stock}, 
the target class $y_{t}$ was found by:
\begin{equation}
 y_{t} = 
  \begin{cases}
   1, & p_{t} \geq p_{t-1}\\
   0, & p_{t} < p_{t-1}
  \end{cases}
\end{equation}

Just like Du and Tanaka-Ishii \cite{du-tanaka-ishii-2020-stock}, we treat the problem as a binary classification problem. For the time window $[t-s+1,...,t-1,t]$ where $s$ is the window size, we consider the time window around day $t$ instead of $t-1$. Having the input window $[t-s+1,..,t-1]$ in the model to output the target class $t$ would be prediction, as the model would not use the articles published on day $t$ to predict the movement at that same day. The prediction problem has been considered very hard by many studies (\cite{famajstorEMH} \cite{TIMMERMANN200415} including \cite{du-tanaka-ishii-2020-stock}) citing the \textit{efficient market hypothesis}. We chose to tackle classification as well due to not having enough timely information to know whether the price or the article happened first \cite{du-tanaka-ishii-2020-stock}. However, we also tested our model with the prediction problem and compared it with classification as it will explained in the section V.


\subsection{Word and Sentence Embeddings}
To get the vectors at the word level, we used the CBOW version of Word2Vec \footnote{We used the gensim library version of Word2Vec. available at: https://radimrehurek.com/gensim/models/word2vec.html} 
due to this version having faster training and slightly better accuracy for frequent words \cite{Mikolov2013EfficientEO}. The model was trained with the data from news article headlines and then, for each headline $i$, the Word2Vec model was used to create word embeddings $w_k$ of dimension 60. All the word embeddings in the headline were further transformed into key vectors $n_{i}^{K}$ of dimension 60, using Term Frequency Inverse Document Frequency score (TFIDF) $\gamma_{k}$
\cite{du-tanaka-ishii-2020-stock}. The key vector is defined by:
\begin{equation}
    n_{i}^{K} = \frac{\sum_{k} \gamma_{k}w_{k}}{\sum_{k} \gamma_{k}}.
\end{equation}
TFIDF is a statistic that gives unique words in a document more importance by computing the number of times the word appears in a document and the number of documents that the word appears in a collection \cite{manning99foundations}

\begin{figure}[H]
    \centering
    \subfloat{\includegraphics[scale=0.25]{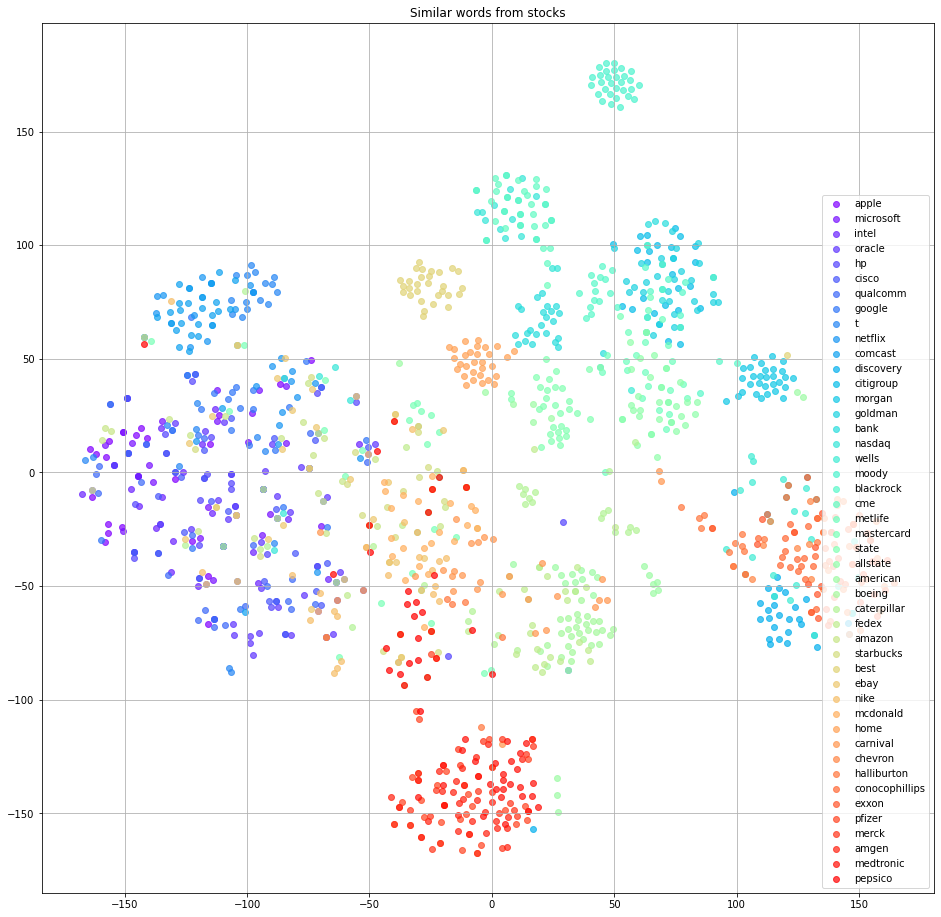}}
    \hfill
    \caption{Visualization of word clusters related to stocks of interest: We ordered the set of companies based on the market sector they belong to and found clusters of similar words to each company using t-SNE \cite{pmlr-v5-maaten09a} to reduce the dimensions of the vectors and visualize them.
The vectors that belong to companies in the same sectors tend to be around the same space, thus strengthening the idea that embeddings can be used to find feature similarities of data.}
    \label{clusters}
\end{figure}
Equation (2) captures the headlines as word-level embeddings. However it is not able to get the context of the headline itself. For example, negation could change the meaning of a sentence with one word. To capture the meaning in a more context sensitive way, we utilized a BERT encoder as a service to output vectors of 1024. The embedding output from the BERT is achieved through the self-attention mechanism in the BERT achieved through transformers \cite{devlin2019bert} \cite{vaswani2017attention}. We then reduced the dimensions to 256 by using principal component analysis (PCA)\cite{du-tanaka-ishii-2020-stock}, thus reducing the number of parameters in the neural network and creating value vectors $n_{i}^{V}$ from the headlines of articles.

Every news article $n_{i}$ was then transformed into a pair of vectors ($n_{i}^{K}, n_{i}^{V}$), utilizing Word2Vec for the word-level ``Key" vector $n_{i}^{K}$, and BERT for the context-level ``Value" vector $n_{i}^{V}$. To save computation time, these were computed once and stored in a dataset where, for each day $t$, there is a pair of sets of key/value vectors $N_{t}^{K}=\{n_{i}^{K}\}_{t}$ and $N_{t}^{V}=\{n_{i}^{V}\}_{t}$ respectively.

The trained stock embeddings were given to us by the authors of the original paper\cite{du-tanaka-ishii-2020-stock}. Let each stock embedding be $s_{j}$ where $j = 1,2,...,J$, and $J=50$ (number of stocks picked from R\&B dataset). We utilized them to compute the \textit{market vector}.
We utilized dot product as the inner product in order to recreate the results from the previous article which aimed to demonstrate the basic capabilities of the stock embedding\cite{du-tanaka-ishii-2020-stock}. 

This was done using the \textit{text feature distiller} process\cite{du-tanaka-ishii-2020-stock}. It starts by calculating the attention score for every article $n_{i}$ of day $t$ for stock $s_j$
\begin{equation}
    score_{i,j} = n_{i}^{K}\cdot s_{j}
\end{equation}
Then the weight of each article $i$ with respect to stock $j$ is found by:
\begin{equation}
    \alpha_{i}^{j} = Softmax(score_{i,j}),
\end{equation}
then finally, the \textit{market vector} $m_{t}^{j}$ for day $t$ with respect to stock $j$ is computed by:
\begin{equation}
    m_{t}^{j}= \sum_{n_{i}^{V}\in N_{t}^{V}}\alpha_{i}^{j}n_{i}^{V}
\end{equation}
The \textit{market vector} is computed for every trading day in the window $s$ with respect to stock $j$. Then we created a sequence of vectors, $M_{[t-s+1,t]}^{j} = [m_{t-s+1}^{j}, m_{t-s+2}^{j}, ..., m_{t-2}^{j}, m_{t-1}^{j}, m_{t}^{j}]$ which is used as the input of the price movement classifier with the target class $y_t$ for day $t$. For this problem, we found the best value for the window $s$ to be 5. The distribution of the correct labels for the five day sequence between an upward stock price movement and downward stock price movement was fairly even ($55\%$ and $45\%$ respectively). 

\subsection{Model Overview}
The proposed model takes as input a sequence of market vectors, $M_{[t-5,t]}^{j} = [m_{t-5}^{j}, m_{t-4}^{j},..., m_{t-1}^{j}, m_{t}^{j}]$ with respect to stock $j$ that are used to classify the price movement\cite{du-tanaka-ishii-2020-stock}. The model then outputs the classification value $\hat{y}_t$. 
The model has a \textit{bidirectional Gated Recurrent Unit}, Bi-GRU layer, which takes in $M_{[t-5,t]}^{j}$ and an initial hidden state $h_{0}$, and uses them to construct a bi-directional encoded vector $h_{t}^{O}$:
\begin{equation}
 h_{t}^{O} = GRU(M_{[t-5,t]}^{j}, h_{0})
\end{equation}  
where $t$ is the trading day we want to classify. The output of the GRU, $h_{t}^{O}$, is then input into a batch normalization layer \cite{ioffe2015batch} and a dropout layer.
After the dropout layer, the classifier estimates the probability by:
\begin{equation}
    \hat{y}_{t}^{j} = softmax(MLP(dropout(batchnorm(h_{t}^{O}))))
\end{equation}
where MLP is a fully connected layer used to predict the binary classification\cite{du-tanaka-ishii-2020-stock}. The model then utilizes cross-entropy loss between target class $y_t^j$ and estimate $\hat{y}_t^j$ for stock $j$, described by:

\begin{equation}
    l_j = -\frac{1}{T} \sum_{t=1}^{T}(y_t^j log(\hat{y}_t^j) + (1-y_t^j)log(1-\hat{y}_t^j)),
\end{equation}
where T is the total number days considered. The mean of all stock losses $j$ is the loss, given by $l=(\sum_{j=1}^{J}l_j)/J$, where $J$ is the total number of stocks considered.

Just like  in Du et al.\cite{du-tanaka-ishii-2020-stock}, a problem with stock movement classification is that single stock classification does not provide enough data to achieve a good performance. In order to address this, a classifier is trained across all the stocks~\cite{du-tanaka-ishii-2020-stock}. In this approach, all stocks share one classifier. This allows for a more generalized model which avoids the overfitting issue with small sample sizes.

\begin{figure}[H]
    \centering
    \includegraphics[scale=0.4250]{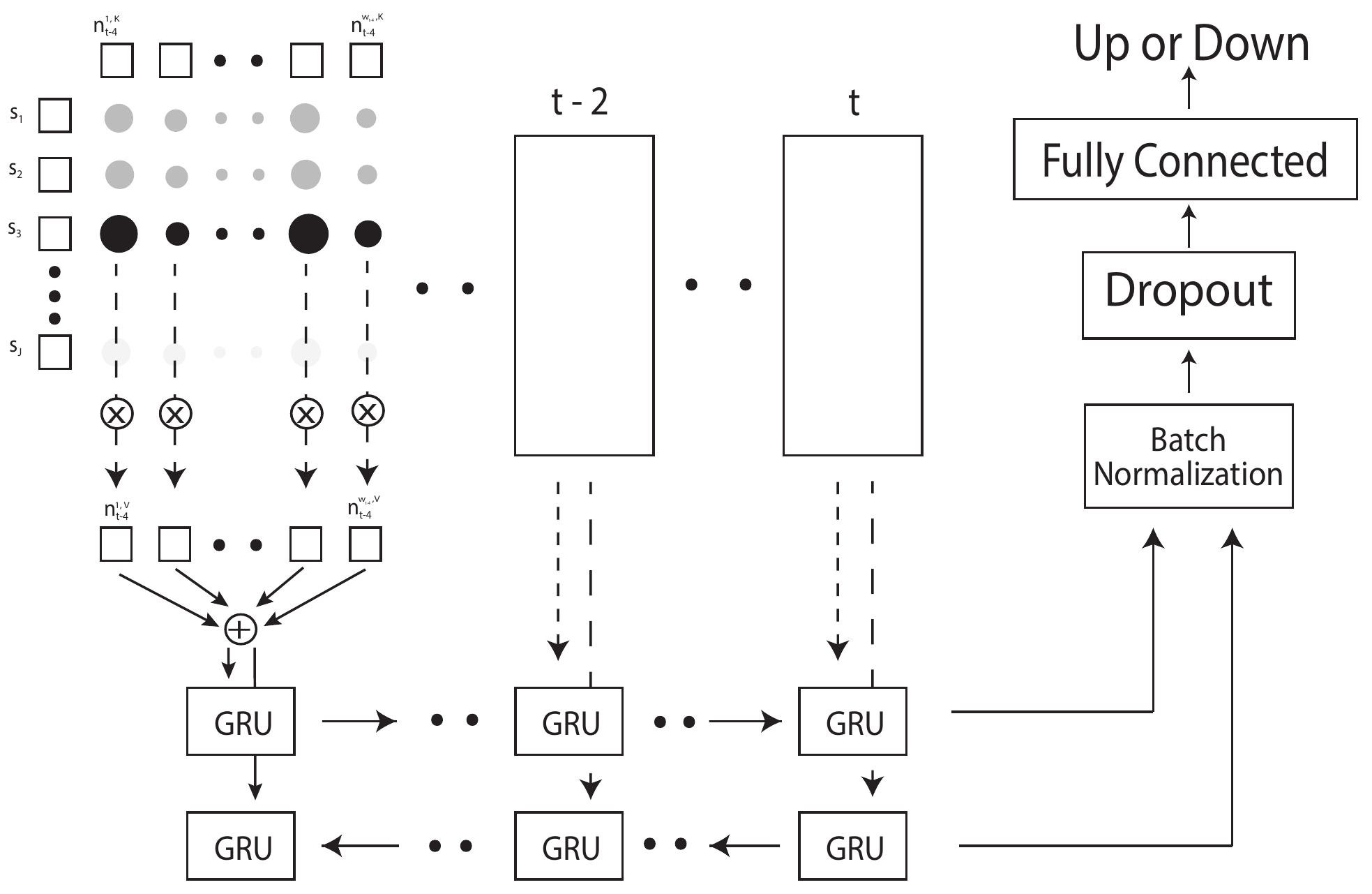}
    \caption{The system diagram of the proposed deep learning model assuming $t-4$ to $t$ are the 
    trading days. The solid circles of different sizes indicate
    different attention weights.}
    \label{Sysdia}
\end{figure}

\subsection{Training Details}
The model was trained in two phases with distinct learning rates: an \textit{Exploration Phase} and an \textit{Exploitation Phase}. We call this method a \textit{Dual-Phase Training}. 

\begin{figure}
    \centering
    \subfloat{\includegraphics[scale=0.5]{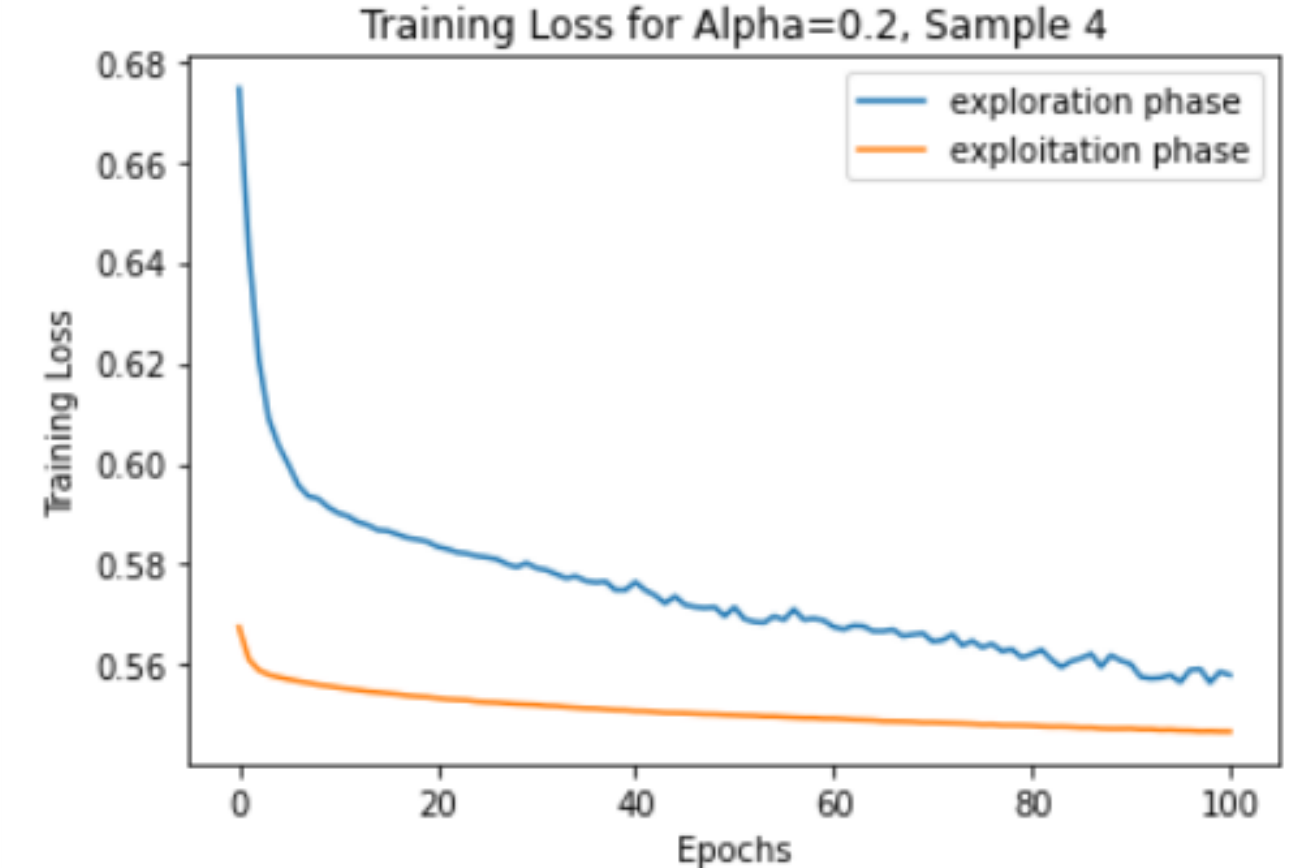}}
    \hfill
    \subfloat{\includegraphics[scale=0.63]{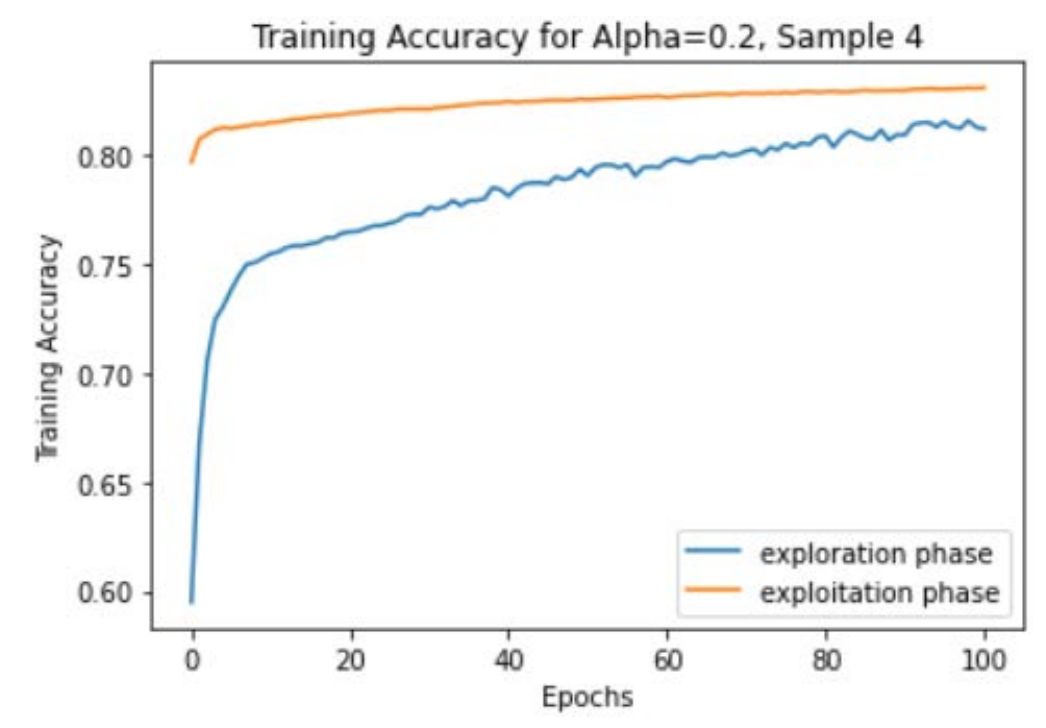}}
    \hfill
    \subfloat{\includegraphics[scale=0.51]{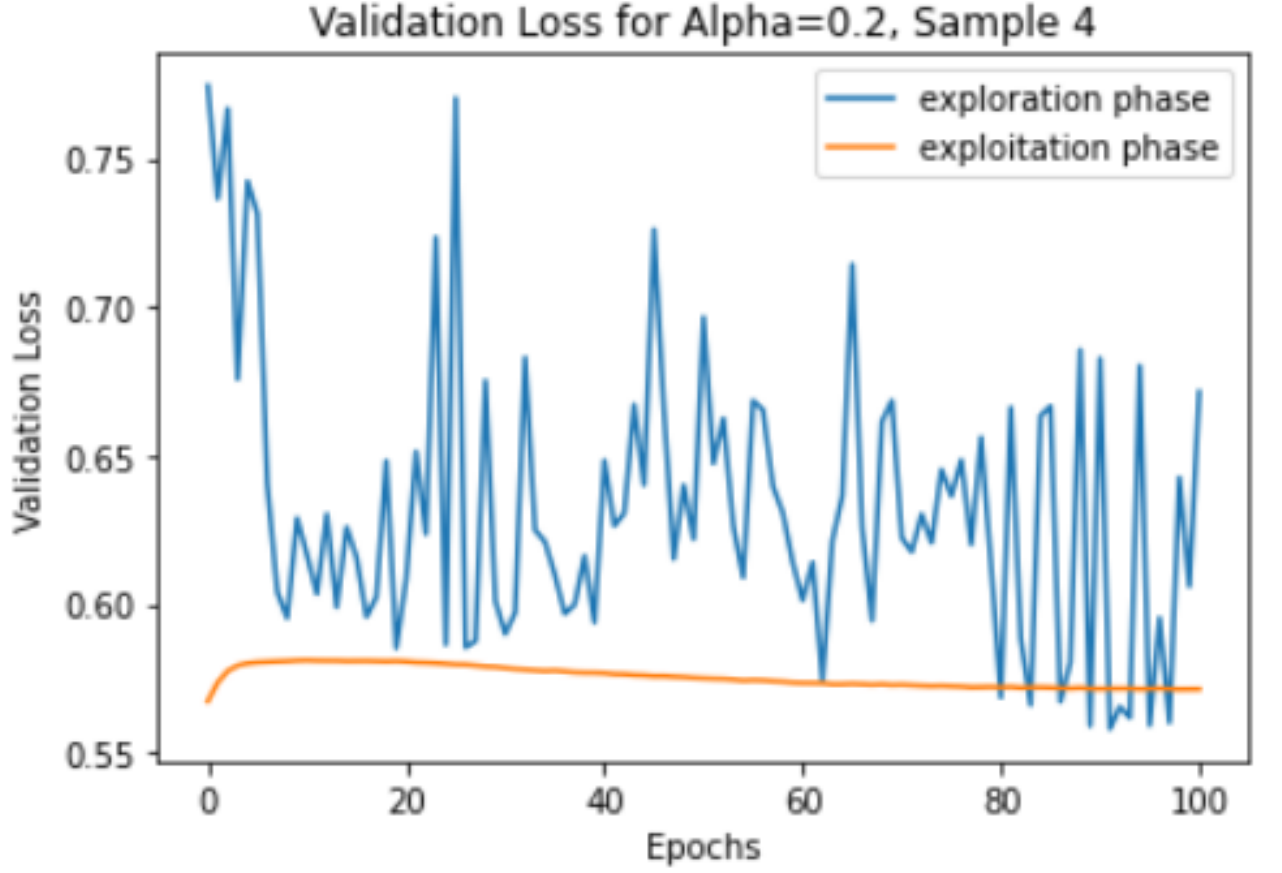}}
    \hfill
    \subfloat{\includegraphics[scale=0.51]{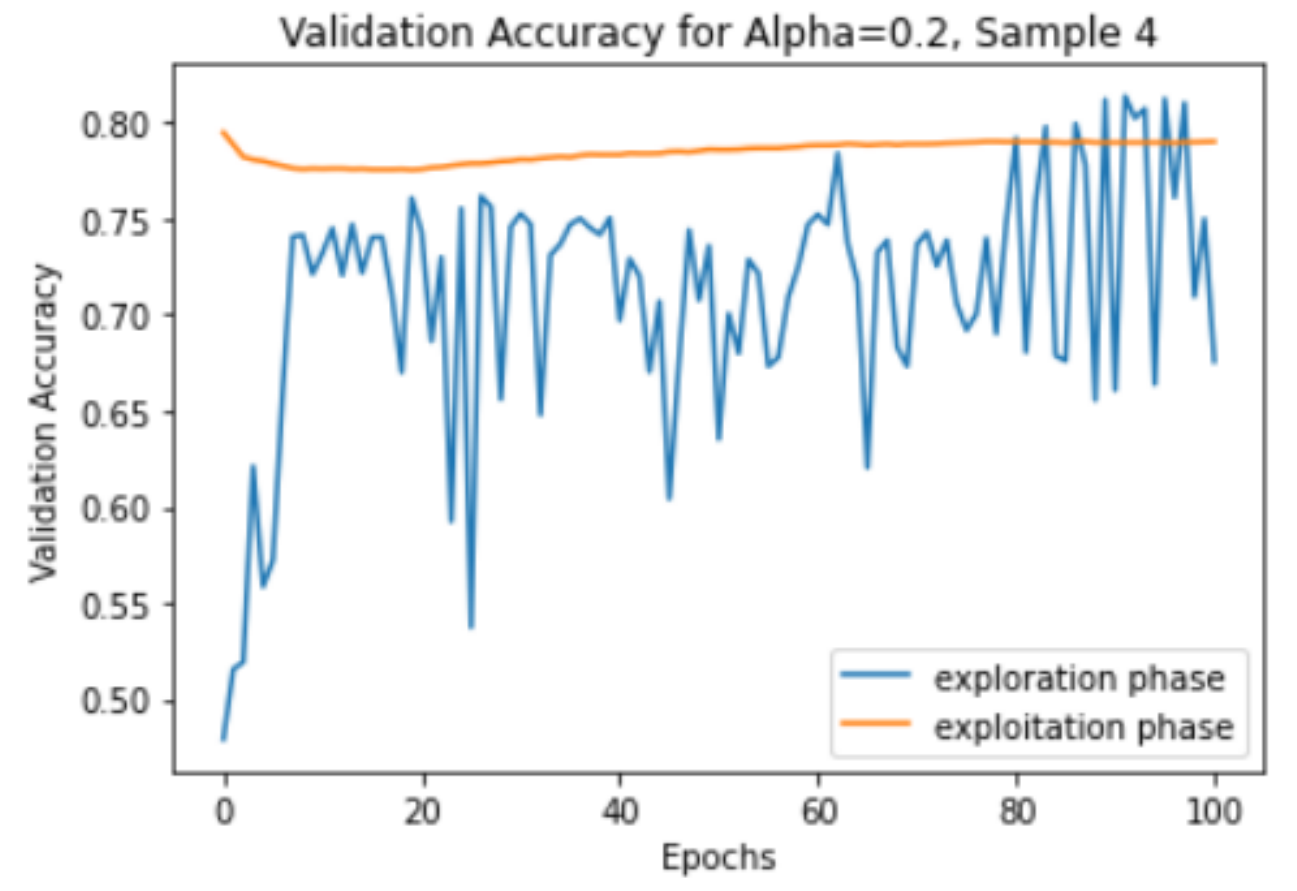}}
    \vspace{0.50in}
    \caption{Example Loss and Accuracy for Training and Validation Sets for $\alpha$=0.2}
    \label{lossaccFullTraining}
\end{figure}

The exploration phase used a learning-rate of $0.0001$, while the exploitation phase used a learning-rate of $0.000001$. The model was trained for 100 epochs in the exploration phase. Then, model with the best validation set performance was trained for an additional 100 epochs in the exploitation phase. Figure \ref{lossaccFullTraining} shows an example of the model's performance across the epochs during both phases.

A mini-batch of 64 samples was used during both training and validation. The dropout was set to 0.2. Since this is a classification problem into either ``stock price increase" or ``stock price decrease" with softmax output, cross-entropy was used as the loss function. The Adam optimizer was used to train the model with weight decay equal to $0.000001$ and the learning rates as above.

\subsection{Label Smoothing}
When dealing with a supervised learning problem with a softmax output, the labels are usually one-hot encoded as vectors of 0's and 1, with 1 indicating the correct label. The cross-entropy loss between the output and correct label vectors is then minimized. 

As Szegedy et al. \cite{szegedy2016rethinking} initially showed, label smoothing regularization can be used to significantly improve model performance during classification, and has since gained widespread popularity. Label smoothing is a technique whereby the one-hot encoded labels are adjusted by $y^{LS}_k = y_k(1-\alpha) + \alpha/K$ for the $k$-th class when calculating the cross-entropy loss\cite{muller2019does}. Here, $K$ is the number of classes, $y_k$ and $y^{LS}_k$ are the labels and modified labels respectively, and $\alpha$ is the label smoothing parameter. In our case, $K=2$ since the two classes are either a distinct increase in stock price or a distinct decrease in stock price. 

\section{Experimental Results}

\subsection{The Dataset}
We used the \textit{Reuters \& Bloomberg} (R\&B) dataset\footnote{Dataset was made open source in Ding et al.\cite{du-tanaka-ishii-2020-stock}, available from  https://github.com/WenchenLi/news-title-stock-prediction-pytorch.} 
for our headline vectors. This dataset contains 552,909 articles combined for a total of 2,605 days or 1,794 trading days. We gathered the price data using Yahoo Finance for the 50 stocks that were mentioned in 100 or more different news articles~\cite{du-tanaka-ishii-2020-stock}. We then created a set of trading days per sample, $N_{[t-s+1,t]}$, where $s$ is the size of the time window. Each sample consists of a tuple $(N_{[t-s+1,t]},y_{t})$.

The original paper proposed using a threshold to calculate ambiguity in a sample using a the log-return between two consecutive days \cite{du-tanaka-ishii-2020-stock}. We did this calculation for $[t-1, t]$ in each sample. After discarding the samples with ambiguity, the total number of samples was 32,204. These samples are used to access the vectors of each article per day and compute the market vector with the stock embeddings provided by Du and Tanaka-Ishii\cite{du-tanaka-ishii-2020-stock}.

For each iteration of training, the dataset was randomly split into $60\%$ training, $20\%$ validation, and $20\%$ test. This was done using the``sklearn" function ``train\_test\_split" with the ``shuffle" parameter set to ``True." Hence, for each new iteration of training, the training, test, and validation sets all may have had different five-day windows as samples compared to previous training iterations.

\subsection{Evaluation Metrics}
We used the standard measurement of accuracy (Acc) and Matthews Correlation Coefficient (MCC), as other studies have used this metric to evaluate stock prediction \cite{xu2018stock} \cite{ding2015Deep}. 

Acc is defined by:
\begin{equation}
    Acc = \frac{Number of Correct Predictions}{Total Number of Predictions Made}
\end{equation}
MCC avoids bias from skewness by measuring the number of correctly classified samples over the total number of samples using a confusion matrix containing the number of samples classified as true positive (tp), false positive (fp), true negative (np), and false negative (fn). MCC is calculated as:
\begin{equation}
    MCC = \frac{(tp*tn)-(fp*fn)}{\sqrt{(tp+fp)(tp+fn)(tn+fp)(tn+fn)}}
\end{equation}

\subsection{Classification Results}

The evaluation of the model was done by averaging the results of ten runs where the dataset was split randomly for each run, as described in Section IV.A. The proposed model achieved an average $80.7\%$ test accuracy and 0.631 average MCC score. This is a dramatic improvement compared with the performance of both Du and Tanaka-Ishii ($68.8\%$)\cite{du-tanaka-ishii-2020-stock} and Ding et al. (Acc: $65.1\%$, MCC: 0.436)\cite{ding2015Deep}, as our model was able to classify over $10\%$ more of the samples correctly. TABLE \ref{tab:modelResults} summarizes the results with test accuracy, standard deviation and MCC score.

\begin{table}[H]
    \centering
    \begin{tabular}{|c||c|c|c|c|c|c|c|}
        \hline
         Model: & Test Accuracy & $\sigma$ & MCC \\
         \hline
         \hline
         Proposed Model & \textbf{0.807} & 0.962 & \textbf{0.631}\\
         Du and Tanaka-Ishii \cite{du-tanaka-ishii-2020-stock} model & 0.688 & 1.67 & - \\
         Ding et al. \cite{ding2015Deep} model & 0.651 & - & 0.436 \\
         \hline
    \end{tabular}
    \caption{Results from test set evaluations comparing proposed model with previous models.}
    \label{tab:modelResults}
\end{table}

\subsection{Label Smoothing Results}
As Szegedy et al.\cite{szegedy2016rethinking} showed, label smoothing regularization can have a significant impact on model performance. Our results mimic this finding. TABLE \ref{tab:labelsmoothingtestresults} shows the test set accuracy when applying different $\alpha$'s to the one-hot encoded labels. The model was trained with the same dual-phase training method described above with the same parameters. Five different run were cared out, with the training, validation, and test sets all shuffled, and the results averaged. 

\begin{table}[H]
    \centering
    \begin{tabular}{|c||c|c|c|c|c|c|c|}
        \hline
         $\alpha$: & 0 & 0.05 & 0.1 & 0.2 & 0.3 & 0.4 & 0.5 \\
         \hline
         \hline
         Run 1 & 0.774 & 0.794 & 0.800& 0.790 & 0.787 & 0.793 & 0.780\\
         Run 2 & 0.768 &0.781 & 0.790 & 0.816 & 0.804 & 0.817 & 0.800\\
         Run 3 &0.781 & 0.780 &0.791 & 0.801 & 0.804 & 0.803 & 0.793\\
         Run 4 &0.742& 0.774 & 0.794 & 0.791 & 0.798 &0.790 & 0.786\\
         Run 5 &0.786 & 0.774 & 0.794 & 0.791 &0.798 & 0.790 & 0.786\\
         \hline
         \hline
         Avg. &0.770 & 0.780 & 0.792 & \textbf{0.804} & 0.798 & 0.803 & 0.789\\
         \hline
    \end{tabular}
    \caption{Accuracies from test set evaluation using label-smoothing with different alphas.}
    \label{tab:labelsmoothingtestresults}
\end{table}

Label smoothing resulted in an increase in the accuracy of the model on the test set by 2-3$\%$ for $\alpha \in [0.1, 0.2, ... 0.4]$. This is an increase of over 10$\%$ compared with the test set accuracy achieved by Du and Tanaka-Ishii (68.8$\%$)\cite{du-tanaka-ishii-2020-stock}. However, as the $\alpha$ increased towards 1, the accuracy began to drop off. This was expected since as $\alpha \to 1$, $y_k^{LS} \to 0.5$ for both classes, making them indistinguishable to the model. 

\subsection{Ablation Study}

We also investigated the impact of several components on the performance of the model by removing them in an ablation study. The label smoothing parameter was set to 0.2 as the model performed the best on the test set with that $\alpha$. The same dual-phase training method as above was used with the same parameters. This was done across five replications with shuffled training, validation, and test sets, and the results averaged.

\begin{table}[H]
    \centering
    \begin{tabular}{|c||c|c|c|c|c|}
    \hline
         Component& None& Weight & Dropout & Batch & Batch Norm \\
         Removed & & Decay & &Norm & + Dropout\\
    \hline\hline
         Run 1 & 0.790 & 0.781 & 0.794 & 0.745 & 0.741\\
         Run 2 & 0.816 & 0.807 & 0.806 & 0.742 & 0.743\\
         Run 3 & 0.801 & 0.811 & 0.800 & 0.740 & 0.745\\
         Run 4 & 0.824 & 0.799 & 0.825 & 0.746 & 0.742\\
         Run 5 & 0.791 & 0.802 & 0.800 & 0.729 & 0.746\\
         \hline
         \hline
         Avg.: & 0.804 & 0.800 & 0.805 & 0.744 & 0.743\\
         \hline
    \end{tabular}
    \caption{Accuracies from test set evaluation after removing selected component(s) with alpha=0.2.}
    \label{tab:AblationTestResults}
\end{table}

As TABLE \ref{tab:AblationTestResults} illustrates, setting both the weight decay and dropout to 0.0 did not have much impact on the model's ability to predict stock price movement for the test set. In fact, removing the dropout actually cased a marginal increase in performance. However, removing the batch normalization layer from the model caused the test set accuracy to drop drastically by about $6\%$. Removing both the batch normalization layer and dropout together had approximately the same effect as removing only the batch normalization layer.

\section{Discussion}
\subsection{Stock Embeddings Relation to Data}
A valid concern for this problem is that the trained stock embeddings provided by Du and Tanaka-Ishii \cite{du-tanaka-ishii-2020-stock} were trained with the same data that we used to train our own model. In order to address this, we introduced a random noise value to the stock embeddings whose scale was one fifth the standard deviation of the embeddings and trained the model with a dataset created from this new embeddings with the dual phase training method and same parameters. The noise model accuracy was 81.1\%  with an MCC score of 0.639.

\begin{table}[H]
    \centering
    \begin{tabular}{|c||c|c|c|c|c|c|c|}
        \hline
         Model: & Test Accuracy & MCC \\
         \hline
         \hline
         Proposed Model & 0.807 & 0.631\\
         Noise Model & \textbf{0.811} & \textbf{0.639} \\
         \hline
         \hline
         Diff. & 0.004 & 0.008 \\
         \hline
    \end{tabular}
    \caption{Results from test set evaluation comparing proposed model with noise model.}
    \label{tab:noiseResults}
\end{table}

\begin{figure}[H]
    \centering
    \subfloat{\includegraphics[scale=0.5]{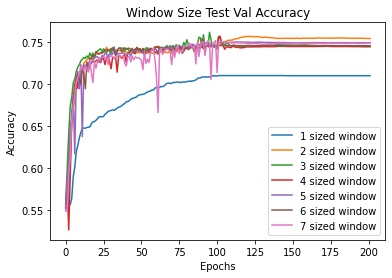}}
    \caption{Validation Accuracy on different window sizes without Batch Normalization}
    \label{fig:valaccWindowSizes}
\end{figure}

To test the dependency of the data, we also checked different window sizes as shown in Figure \ref{fig:valaccWindowSizes}. These results show that our system is not sensitive to the number of days as long as there is a context. Still, a window of size one was able to predict better than the previous methods. Because we used a Bi-GRU, the backwards GRU does not change with respect to the size of the window. However, the system managed to learn effective features for a window of size 2 or more. 

\subsection{Validation Set Fluctuations}
\begin{figure}[H]
    \centering
    \subfloat{\includegraphics[scale=0.55]{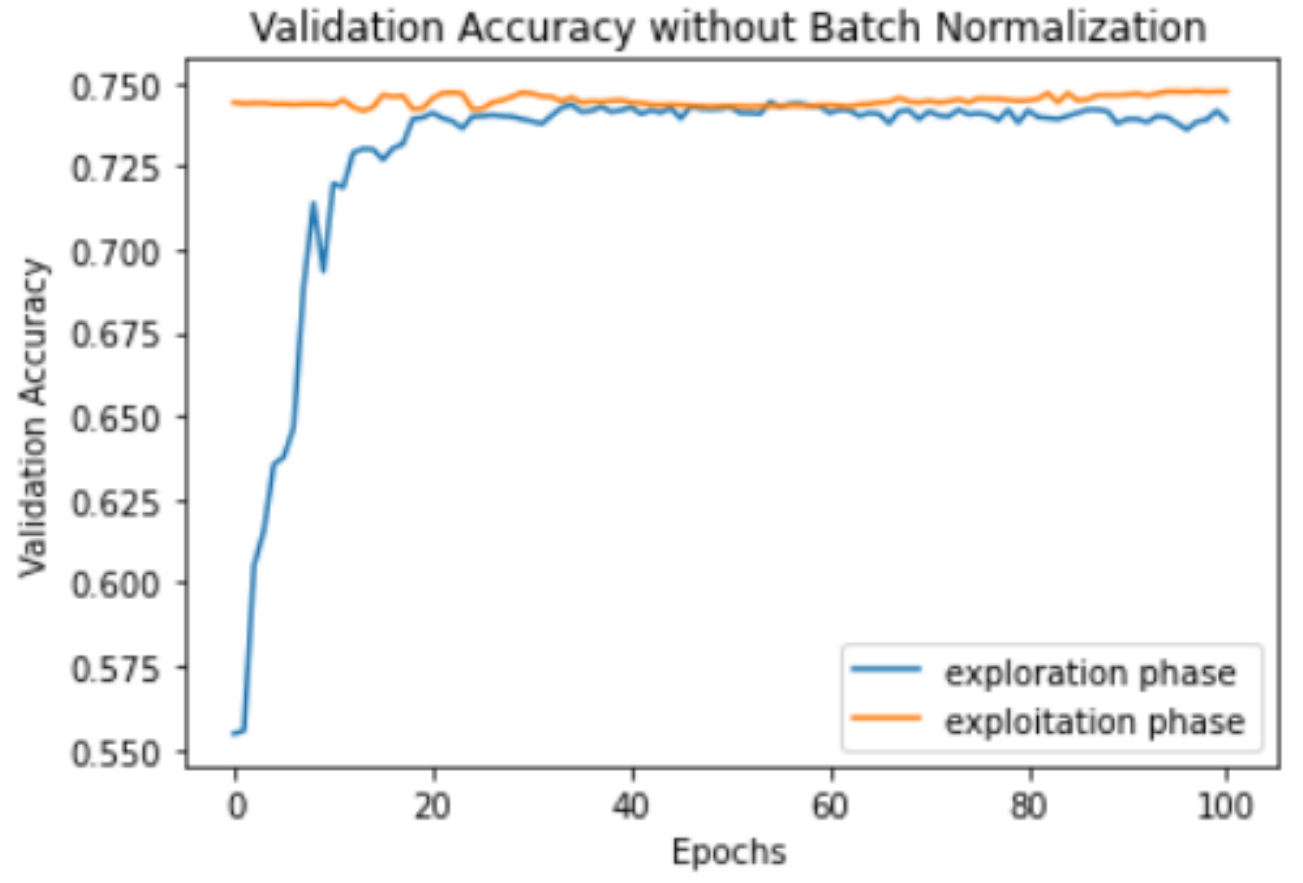}}
    \caption{Validation Accuracy for $\alpha=0.2$ with Batch Normalization Removed}
    \label{fig:valaccNOBATCHNORM}
\end{figure}

\begin{figure}[H]
    \centering
    \includegraphics[scale=0.55]{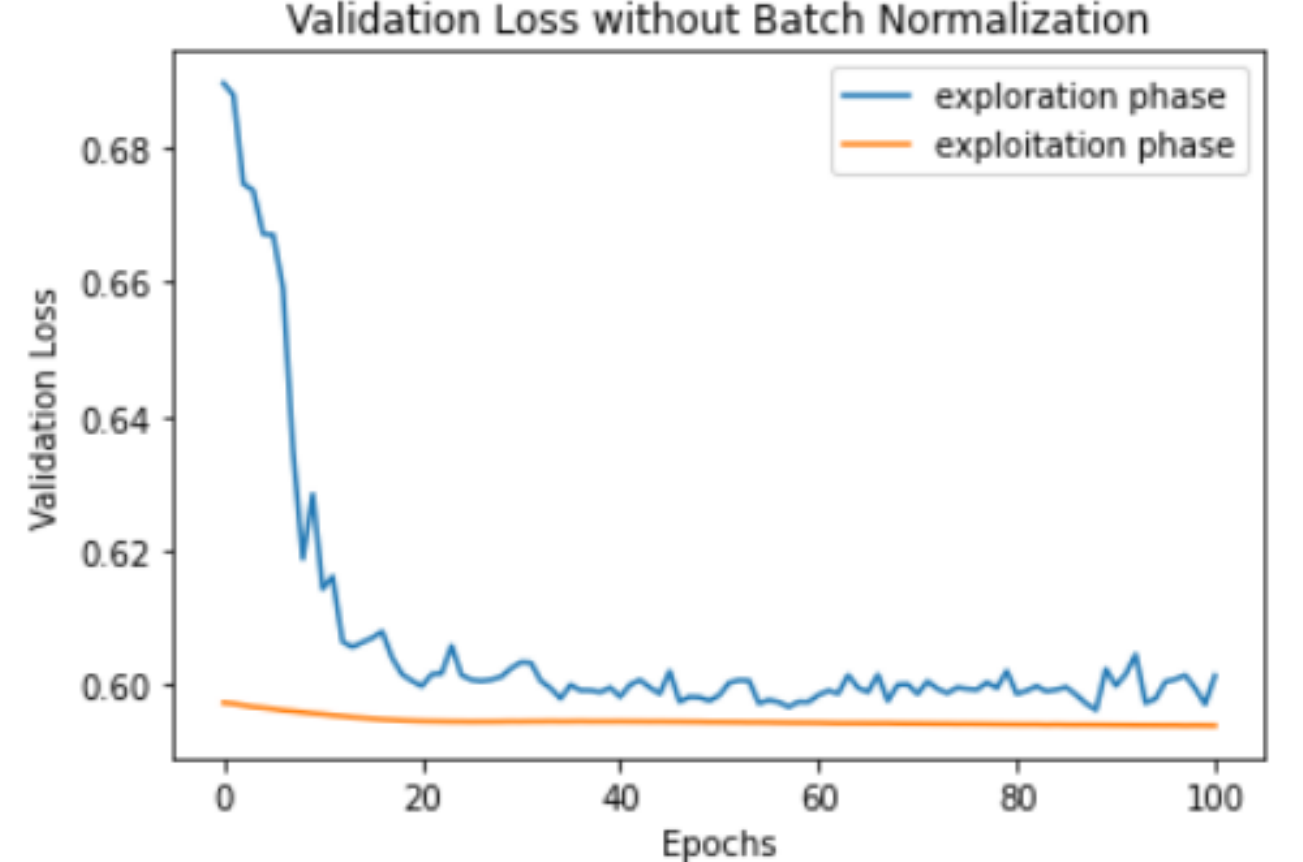}
    \caption{Validation Loss for $\alpha=0.2$ with Batch Normalization Removed}
    \label{fig:vallossBATCHNORM}
\end{figure}

It is clear from Figure \ref{lossaccFullTraining} that during the exploration phase when the learning rate is high (0.0001), the accuracy and loss on the validation set fluctuate wildly. However, if we examine the validation accuracy and loss when the batch normalization is removed, they become more stable, as seen in Figures \ref{fig:valaccNOBATCHNORM} and \ref{fig:vallossBATCHNORM}.


We believe this may be related to the internal statistics that batch normalization tracks across each epoch, but more investigation is necessary to confirm this. However, it is clear from TABLE \ref{tab:AblationTestResults} that performing batch normalization significantly increases the model's accuracy on the test set. This indicates that batch normalization is indeed having the intended effect of improving the model's generalization\cite{ioffe2015batch}.

\subsection{Prediction Vs Classification}

In order to test whether the model could predict price movement in the near future, we trained our proposed classification model with a prediction dataset. This involves using the market vectors in the time range $[t-s+1,t-1]$ and predicting the price movement for day $t$. We call this the prediction dataset. To be able to prove the model is able to predict, we tested as follows:

\begin{table}[H]
    \centering
    \begin{tabular}{|c||c|c|c|c|c|c|c|}
        \hline
         Model: & Test Accuracy & MCC\\
         \hline
         \hline
         Prediction model with prediction set & 0.765 & 0.557\\
         Classification model with classification set & 0.817 & 0.644\\
         Prediction model with classification set & 0.453 & 0.407\\
         Classification model with prediction set & 0.452 & 0.496\\
         \hline
    \end{tabular}
    \caption{Results from test set evaluations comparing prediction with classification.}
    \label{tab:predclassresults}
\end{table}

\begin{figure}[H]
    \centering
    \subfloat{\includegraphics[scale=0.45]{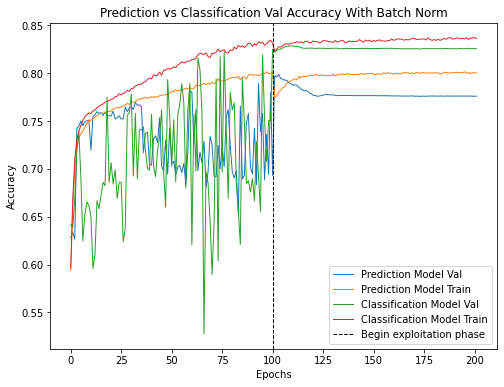}}
    \hfill
    \caption{The plots show that the proposed model is able to do prediction as well as classification}
    \label{predclasscurves}
\end{figure}

Table \ref{tab:predclassresults} shows how we used the prediction dataset and the classification dataset on both models to illustrate how not using the correct time range for the trained task can lead to poor results. The classification model cannot classify using the prediction set and inversely, the prediction model cannot predict using the classification set. Figure \ref{predclasscurves} shows the accuracy of both models during training for this test.

\subsection{Implications of Efficient Market Hypothesis (EMH)}
Perhaps the study is inconsistent with the Efficient Market Hypothesis (EMH), which states that prices in an efficient market fully reflect all available, relevant information\cite{famajstorEMH}. Once patterns are discovered, they provide an advantage only for a limited time, as rational traders adjust their behavior to compensate for this new information, the pattern will be destroyed\cite{TIMMERMANN200415}.


However, in recent years, there has been much criticism of the EMH from proponents of behavioral finance\cite{malkiel2003efficient}. One of the foundational assumptions of economics is that individuals behave rationally when exposed to incentives. This is clearly not always the case and is the basis of behavioral finance\cite{RITTER2003429}. Therefore, since the news and other new information does have an effect on the stock price, this may place the market in so-called ``over-reaction and under-reaction states," which may be taken advantage of by knowledgeable investors\cite{behavioralfinpaper1}. 
It is also logical to assume that when relationships between economic forces are more complicated, such as between the news and stocks, different economic actors may interpret those patterns differently, or fail to recognize them at all. Indeed, problems of incomplete information have been studied by economists for decades\cite{merton1987simple}, and the presence of complete information is another assumption at the foundation of classical economics models such as the EMH \cite{famajstorEMH}. Hence, since this assumption is not incredibly realistic, if our model were to be able to uncover more information about how the news and stocks are related than Du and Tanaka-Ishii\cite{du-tanaka-ishii-2020-stock} through different regularization techniques, our model could provide additional financial advantage. 


\section{Conclusion}

By modifying the neural network architecture and introducing further generalization techniques, we were able to significantly improve the performance of the model first introduced by Du and Tanaka-Ishii \cite{du-tanaka-ishii-2020-stock}, from 68.8\% to 79.3\% test accuracy. Smoothing the one-hot encoded labels during loss calculation further increased performance to a maximum accuracy of 80.4\%. An ablation study was able to confirm the significant impact that batch normalization had on model performance, while the effects of introducing weight decay and a dropout layer were much less pronounced. 

Since we utilized the same embeddings that were trained by Du and Tanaka-Ishii's model, it may be possible to increase the test set accuracy even further by implementing a joint training of the stock and news embeddings together with the network. 

We also foresee that the accuracy could be increased by making the classification problem more realistic. One simple way to do this would be to expand the classes from ``up" and ``down" to ``up", ``down", and ``no change". This would involve reincorporating the stocks with little price movements back into the dataset that were originally removed by Du and Tanaka-Ishii \cite{du-tanaka-ishii-2020-stock}. 

Additionally, abandoning one-hot encoding altogether in favor of \textit{Soft Label Assignment} (SLA) might lead to further accuracy improvements. As first described by Alishahrani et al. \cite{alshahrani2021optimism}, SLA in this case would involve estimating the probability distribution of the stock price movement classifiers for each stock. Then the stocks would be assigned probabilities for each classifier for a given window during the calculation of the loss. 



\bibliographystyle{IEEEtran}
\bibliography{merged}

\end{document}